\documentclass[11pt]{article}
\usepackage{coling2020}
\usepackage{tikzsymbols}
\usepackage{amsfonts}
\usepackage{multirow}
\usepackage{times}
\usepackage{url}
\usepackage{latexsym}

\colingfinalcopy 

\title{Want to Identify, Extract and Normalize Adverse Drug Reactions in Tweets? Use RoBERTa}

\author{Katikapalli Subramanyam Kalyan \\
  Department of Computer Applications \\
  NIT Trichy, India \\
  {\tt kalyan.ks@yahoo.com} \\\And
  S.Sangeetha \\
  Department of Computer Applications \\
  NIT Trichy, India \\
  {\tt sangeetha@nitt.edu} \\}

\date{}

\begin{document}
\maketitle
\begin{abstract}
  This paper presents our approach for task 2 and task 3 of Social Media Mining for Health (SMM4H) 2020 shared tasks. In task 2, we have to differentiate adverse drug reaction (ADR) tweets from nonADR tweets and is treated as binary classification. Task3 involves extracting ADR mentions and then mapping them to MedDRA codes. Extracting ADR mentions is treated as sequence labeling and normalizing ADR mentions is treated as multi-class classification.   Our system is based on pre-trained language model RoBERTa and it achieves a) F1-score of 58\% in task2 which is 12\% more than the average score b) relaxed F1-score of 70.1\% in ADR extraction of task 3 which is 13.7\% more than the average score and relaxed F1-score of 35\% in ADR extraction + normalization of task3 which is 5.8\% more than the average score. Overall, our models achieve promising results in both the tasks with significant improvements over average scores.
\end{abstract}

\section{Introduction}

\blfootnote{
    %
    %
    \hspace{-0.65cm}  
    ***** preprint under review ******
    %
    %
    %
    %
}

Social media platforms in particular, twitter is used extensively by common public to share their experiences which also includes health-related information like ADRs they experience while consuming drugs. Adverse Drug Reaction (ADR) refers to unwanted harmful effect following the use of one or more drugs. The abundant health-related social media data can be utilized to enhance the quality of services in health-related applications \cite{kalyan2020secnlp}. Our team participated in task 2 and task 3 of SMM4H 2020 shared task. Task 2 aims at identifying whether a tweet contains ADR mention or not. Task 3 aims at extracting ADR mentions and then normalizing them to MedDRA concepts. Our team achieved a) F1-score of 58\% in task 2 and b) relaxed F1-score of 35\% in task 3. Overall, our models achieve promising results in both the tasks with significant improvements over average scores.

\section{Task 2 – Identification of ADR tweets}

\subsection{Problem Definition and Dataset}
Task2 aims at identifying whether a tweet contains ADR mention or not. An example of an ADR tweet is:
`\textit{thank god for vyvanse \#addicted}`. Here `\textit{addicted}` is the adverse drug reaction that happened because of consumption of the drug `\textit{vyvanse}`. An example of a nonADR tweet is  `\textit{never take paxil \#js}`.  In this task, we learn a classification model which outputs the label 1 or 0 for a given tweet depending on whether it contains ADR mentions or not.  In this dataset, the training set consists of 20544 tweets (18641 nonADR and 1903 ADR tweets), validation set consists of 5134 tweets (4660 nonADR and 474 ADR tweets) and the test set consists of 4759 tweets. 

\subsection{Methodology}

\subsubsection*{Pre-processing}
We apply the following pre-processing steps
\begin{itemize}
    \item Lowercase the text and remove consecutively repeating characters in the words (e.g., feeeeel $\to$ feel).
    \item Remove urls, @user mentions, retweet tag (rt), non-ASCII and punctuation characters.
    \item Expand English contractions (e.g., can’t $\to$ cannot) and replace interjections with their meanings (e.g., ouch, oww $\to$ pain).
    \item Replace character smiley (e.g., :) $\to$ happy) and emoji (e.g., \dSmiley $\to$ grinning face) with their text descriptions. 
\end{itemize}

\subsubsection*{Model Description}
In recent times, the evolution of pretrained language models like BERT \cite{devlin2019bert}, RoBERTa \cite{liu2019roberta} changed the scenario in natural language processing. These models learn universal language representations from large training corpus and they can be used in downstream tasks by adding one or two layers which are specific to the task \cite{qiu2020pre}.  Our model is based on RoBERTa. We add task-specific sigmoid layer on the top of RoBERTa and then fine-tune the entire model using training dataset.  We consider the final hidden state vector $t_{<s>} \in \mathbb{R}^h$ of the special token \textless s\textgreater as tweet representation. Here $h$ represents hidden state vector size in RoBERTa-base and it is equal to 768. The vector $t_{<s>}$  is passed through sigmoid layer to get the prediction $\hat{y}$. Overall, the label $\hat{y}$ is computed as :
 \begin{equation}
     t_{<s>} = RoBERTa(tweet)
 \end{equation}
 \begin{equation}
     \hat{y} = Sigmoid(W^Tt_{<s>} + b)
 \end{equation}
 
\noindent Here $W \in \mathbb{R}^{h\times1}$ and $ b \in \mathbb{R}$ are learnable parameters of sigmoid layer.

\subsection{Experiments and Results}
As ADR tweets are less in number compared to nonADR tweets in the training set, we augment training set with ADR tweets from SMM4H 2017 \cite{sarker2018data} and SMM4H 2019 \cite{weissenbacher2019overview} ADR tweets classification datasets. Further, we include only randomly chosen 90\% of nonADR tweets in the training set. By conducting random search over the range of hyperparameters values, we arrive at the following optimal set of hyperparameter values: batch size = 128, learning rate = 3e-5, dropout = 0.2 (applied on $t_{<s>}$ vector to reduce overfitting) and epochs = 10. We implement our model in PyTorch framework using transformers library from huggingface \cite{wolf2019huggingface}.  

\begin{table*}[h]
\begin{center}
\begin{tabular}{|l|l|l|l|}
\hline \footnotesize  Model &  \footnotesize Precision &  \footnotesize Recall	&  \footnotesize F1 \\ \hline
\footnotesize Roberta-base(ours)  & \footnotesize	52.00 &\footnotesize 65.00 & \footnotesize	\textbf{58.00} \\
\footnotesize Average scores  &	\footnotesize 42.00 &\footnotesize	59.00 & \footnotesize	46.00 \\
\hline
 \end{tabular}
\end{center}
\caption{\label{results-task2} Task 2 - ADR Tweet classification results on test data} 
\end{table*}
We report performance of our model and average scores in task2 – ADR tweets classification in Table \ref{results-task2}. Our model achieves an F1-score of 58\% and it is 12\% more than the average score. 

\section{Task 3 – Extract and Normalize ADR Mentions}
\subsection{Problem Definition and Dataset }
This task involves ADR extraction followed by normalization. In the first part (ADR Extraction), to extract ADR mentions the model has to identify ADR tweets and then extract ADR mentions by identifying the spans in tweets. A tweet can have more than one ADR mention also and an ADR mention can be a sequence of words also. Example of a tweet with ADR mentions

\textit{@coolpharmgreg i don't care if they are toxic haha putting the cipro drops in is essentially equivalent to torture \#oww}

\noindent Here `\textit{oww}’, `\textit{toxic}’ and `\textit{equivalent to torture}’ are the adverse drug reactions due to the consumption of the drug `\textit{cipro}`. In the second part (ADR normalization), the extracted ADR mentions are mapped to the standard concepts in MedDRA vocabulary. In the above example, the ADR mentions `\textit{oww}’, `\textit{toxic}’, and `\textit{equivalent to torture}’ are mapped to the concepts  `\textit{pain (10033371)}’, `\textit{drug toxicity (10013746)}’ and `\textit{feeling unwell (10016370)}’ respectively.The dataset for this task consists of training set with 1862 tweets (1080 ADR tweets with 1464 ADR mentions and 782 nonADR tweets), validation set with 428 tweets (233 ADR tweets with 365 ADR mentions and 195 nonADR tweets) and test set with 976 tweets. 

\subsection{Methodology}
ADR extraction is viewed as sequence labeling which is nothing but assigning a label to each of the tokens in the sequence. We follow BIO tagging: the tag `B-ADR’ represents tokens at the beginning of ADR mention, `I-ADR’ represents tokens inside ADR mention and `O’ represents nonADR tokens. We experiment with two models for this task. a)The first model is based on RoBERTa i.e., RoBERTaForTokenClassification. b)The second model is multi-task learning based RoBERTa. In this, ADR tweet identification is the auxiliary task and ADR extraction is the main task. As these two tasks are similar in nature, by joint learning, the knowledge gained in auxiliary task improves the performance of the main task ADR extraction \cite{caruana1997multitask,crichton2017neural}. 
Following the recent work in normalizing medical concepts \cite{kalyan2020bertmcn,SUBRAMANYAM20201353,kalyan2020medical}, we treat concept normalization as multi-class classification and experiment with RoBERTa. 

\subsection{Experiments and Results}
For ADR extraction, in case of RoBERTa based model, we set batch size = 64, epochs = 20 and learning rate = 0.00003. In case of multitask learning model, $loss = \lambda \times L_{ADE} + (1-\lambda) \times L_{ADR}$. Here $L_{ADE}$  represents loss related to ADR extraction and $L_{ADR}$  is loss related to ADR detection. The value of $\lambda$ is set to 0.8. Here, the model is trained for 30 epochs with learning rate = 3e-5 and batch size = 64. For ADR normalization, we use a learning rate of 3e-5 and batch size of 128. For both ADR extraction and normalization, we use AdamW optimizer \cite{loshchilov2018decoupled}.

\begin{table*}[h]
\begin{center}
\begin{tabular}{|l|l|l|l|l|l|}
\hline
\footnotesize Model & \footnotesize Evaluation  & \footnotesize Type    & \footnotesize  Precision & \footnotesize Recall & \footnotesize F1   \\ \hline
\multirow{4}{*}{\footnotesize RoBERTa}     & \multirow{2}{*}{\footnotesize NER} & \footnotesize Relaxed & \footnotesize 63.0      & \footnotesize 78.9   & \footnotesize \textbf{70.1} \\
                             &                             & \footnotesize Strict  & \footnotesize 41.1      & \footnotesize 54.2   & \footnotesize  46.8 \\ \cline{2-6}
                             & \multirow{2}{*}{\footnotesize NER + Norm} & \footnotesize Relaxed & \footnotesize 30.4      & \footnotesize 39.8   & \footnotesize  34.5 \\
                             &                             & \footnotesize Strict  & \footnotesize 23.6      & \footnotesize 31.1   & \footnotesize  26.8 \\ \hline
\multirow{4}{*}{\footnotesize RoBERTa+MTL} & \multirow{2}{*}{\footnotesize NER}        & \footnotesize Relaxed & \footnotesize 65.1      & \footnotesize 72.8   & \footnotesize 68.7 \\
                             &                             & \footnotesize Strict  & \footnotesize 45.2      & \footnotesize 52.5   & \footnotesize \textbf{48.6} \\ \cline{2-6}
                             & \multirow{2}{*}{\footnotesize NER + Norm} & \footnotesize Relaxed & \footnotesize 32.6      & \footnotesize 37.7   & \footnotesize \textbf{35.0} \\
                             &                             & \footnotesize Strict  & \footnotesize 25.5      & \footnotesize 29.6   & \footnotesize \textbf{27.4} \\ \hline
\multirow{2}{*}{\footnotesize Average  scores}     & \footnotesize NER  & \footnotesize Relaxed & \footnotesize 60.7    & \footnotesize 55.7   & \footnotesize 56.4 \\
                             & \footnotesize NER + Norm & \footnotesize Relaxed & \footnotesize 31.2      & \footnotesize 29.0   & \footnotesize 29.2 \\ \hline
\end{tabular}
\caption{\label{results-task3} Task 3 - ADR Extraction and Normalization results  on test data. Here NER represents ADR Extraction and Norm represents ADR Normalization.} 
\end{center}
\end{table*}
The performance of our models is reported in Table \ref{results-task3}. From the table we observe that, a) RoBERTa based model achieved relaxed F1 score of 70.1\% in ADR extraction which is 13.7\% more than the relaxed average score b)Multi-task learning RoBERTa based model achieved relaxed F1 score of 35\% in NER+Norm which is 5.8\% more than the relaxed average score. 

\section{Conclusion}
In this work, we explored the effectiveness of RoBERTa to identify, extract and normalize ADR mentions in tweets. In both task2 and task3, our proposed models achieved promising results with significant improvements over average scores.

\bibliographystyle{coling}
\bibliography{coling2020}

\end{document}